%% file: main.tex
\documentclass{article}

\usepackage[preprint]{cls/corl_2022} 

\usepackage[table,usenames,dvipsnames]{xcolor}      
\usepackage{enumitem}                               
\usepackage{extarrows}                              
\usepackage[noadjust]{cite}                         
\usepackage{verbatim}

\usepackage{amsmath,amssymb,amsfonts,dsfont} 
\usepackage{amsthm}
\usepackage{bbm}
\usepackage{algorithm,algorithmicx,listings}        
\usepackage[noend]{algpseudocode}             

\usepackage{graphicx,tabularx,subcaption}
\usepackage[export]{adjustbox}
\usepackage{wrapfig}
\usepackage{makecell,booktabs}

\usepackage[font={small}]{caption}
\usepackage[titletoc,title]{appendix}

\newcommand{\prl}[1]{\left(#1\right)}
\newcommand{\brl}[1]{\left[#1\right]}
\newcommand{\crl}[1]{\left\{#1\right\}}

\newcommand{\bcalA}{\bar{\calA}}
\newcommand{\bcalM}{{\bar{\calM}}}
\newcommand{\bbfa}{\bar{\bfa}}
\newcommand{\btheta}{{\bar{\theta}}}
\newcommand{\bpi}{{\bar{\pi}}}

\newcommand{\ubar}[1]{\text{\b{$#1$}}}

\input{cls/sym.tex}

\title{Latent Policies for Adversarial Imitation Learning}

\author{
  Tianyu Wang\\
  Department of Electrical and Computer Engineering\\
  University of California, San Diego,
  United States\\
  \texttt{tiw161@eng.ucsd.edu} \\
  \And
  Nikhil Karnwal\\
  Department of Electrical and Computer Engineering\\
  University of California, San Diego,
  United States\\
  \texttt{nkarnwal@eng.ucsd.edu} \\
  \And
  Nikolay Atanasov \\
  Department of Electrical and Computer Engineering\\
  University of California, San Diego,
  United States\\
  \texttt{natanasov@eng.ucsd.edu} \\
}

\begin{document}
\maketitle


\begin{abstract}
This paper considers learning robot locomotion and manipulation tasks from expert demonstrations. Generative adversarial imitation learning (GAIL) trains a discriminator that distinguishes expert from agent transitions, and in turn use a reward defined by the discriminator output to optimize a policy generator for the agent. This generative adversarial training approach is very powerful but depends on a delicate balance between the discriminator and the generator training. In high-dimensional problems, the discriminator training may easily overfit or exploit associations with task-irrelevant features for transition classification. A key insight of this work is that performing imitation learning in a suitable latent task space makes the training process stable, even in challenging high-dimensional problems. We use an action encoder-decoder model to obtain a low-dimensional latent action space and train a LAtent Policy using Adversarial imitation Learning (LAPAL). The encoder-decoder model can be trained offline from state-action pairs to obtain a task-agnostic latent action representation or online, simultaneously with the discriminator and generator training, to obtain a task-aware latent action representation. We demonstrate that LAPAL training is stable, with near-monotonic performance improvement, and achieves expert performance in most locomotion and manipulation tasks, while a GAIL baseline converges slower and does not achieve expert performance in high-dimensional environments.
\end{abstract}


\input{tex/Introduction.tex}
\input{tex/RelatedWork.tex}
\input{tex/Preliminaries.tex}
\input{tex/LatentAction.tex}
\input{tex/TaskAgnostic.tex}

\input{tex/TaskAware.tex}

\input{tex/Limitations.tex}
\input{tex/Experiments.tex}

\input{tex/Conclusion.tex}



\clearpage
\acknowledgments{We gratefully acknowledge support from ONR SAI N00014-18-1-2828.}


\bibliography{bib/ref}  

\end{document}

%% file: cls/sym.tex
\newcommand{\calA}{{\cal A}}
\newcommand{\calB}{{\cal B}}

\newcommand{\calL}{{\cal L}}
\newcommand{\calM}{{\cal M}}

\newcommand{\calP}{{\cal P}}

\newcommand{\calS}{{\cal S}}

\newcommand{\bfa}{\mathbf{a}}

\newcommand{\bfs}{\mathbf{s}}

\newcommand{\bbE}{\mathbb{E}}

\newcommand{\bbR}{\mathbb{R}}

%% file: tex/Introduction.tex
\section{Introduction}
\label{sec:introduction}

The reward function of a reinforcement learning problem serves a dual role of specifying the task objectives and providing a signal for the value and policy function optimization. It is often too challenging to optimize a policy with a sparse reward or too difficult to design an informative reward function that captures the intricacies of the desired behaviors \citep{Schaal1996LfD, Argall2009Survey, Hadfield2017IRD}. Imitation learning (IL) \citep{Schaal1996LfD, Pastor2009LfD, Hussein2017Imitation, Zhu2018RLIL} overcomes these challenges by mimicking expert behaviors from demonstrations. One common approach in IL is inverse reinforcement learning (IRL) \citep{Ng2000IRL, Abbeel2004IRL, Neu2012Apprenticeship} where the expert's reward function is inferred from demonstrations and a policy is trained subsequently to optimize the learned reward.


Inspired by generative adversarial networks (GANs) \citep{Goodfellow2014GAN}, adversarial imitation learning (AIL) methods \citep{Ho2016GAIL, Fu2017AIRL, Finn2016GAN-GCL} train a discriminator that distinguishes the expert's and agent's transitions, and in turn use a reward defined by the discriminator output to optimize a generator policy. For complex environments with high-dimensional state spaces, training AIL algorithms is challenging because the discriminator may easily overfit \citep{Orsini2021AIL} or exploit spurious associations between task-irrelevant features and class labels, providing little information for the generator learning. Explicit regularization techniques like gradient penalty \citep{Gulrajani2017WGAN} and spectral normalization \citep{Miyato2018Spectral} are applied to alleviate such issues but require careful tuning and their effects are environment dependent \citep{Orsini2021AIL}. Alternatively, discriminator training may be restricted by using only task-relevant features \citep{Zolna2019TRAIL}.

Consider a manipulation task requiring a robot arm to control a gripper (see Figure~\ref{fig:toy_env}). A canonical parametrization of the action space utilizes joint angles, angular velocities or torques. A key insight of our work is that a successful trajectory in the joint configuration space may be complicated but its embedding in the end effector (e.g., fingers or gripper) configuration space is relatively simple. We postulate that learning a policy in the original action space of a robot may be difficult and hard to generalize but restricting the learning process to a suitable latent task space, such as the end effector pose in a manipulation task, makes it easier and more generalizable.


The main contribution of this work is an approach to learn policy functions in latent action space that captures the task structure when imitating expert behavior in complex robotic tasks. Our proposed algorithm, termed \textit{\ubar{L}\ubar{A}tent \ubar{P}olicies for \ubar{A}dversarial Imitation \ubar{L}earning} (LAPAL), trains an action encoder-decoder model, to provide a latent structured action space for efficient adversarial imitation learning. We demonstrate that the action encoder-decoder model can be trained offline with expert demonstrations only to learn latent action representations, and trained online to learn latent representations that align with the imitation learning task objective. Through experiments on locomotion and manipulation tasks, we find that LAPAL can successfully learn latent policies from adversarial imitation training. Especially it exhibits stable and near-monotonic learning, reaching expert performance asymptotically in most tasks while baseline converges slower or could not recover the optimal policy in high-dimensional systems. We further demonstrate the ability to transfer the latent policy to new robot configurations by fine-tuning the action encoder-decoder model without additional environment interactions. 


\begin{table}
\begin{minipage}{0.3\linewidth}
    \centering
    \includegraphics[width=\linewidth]{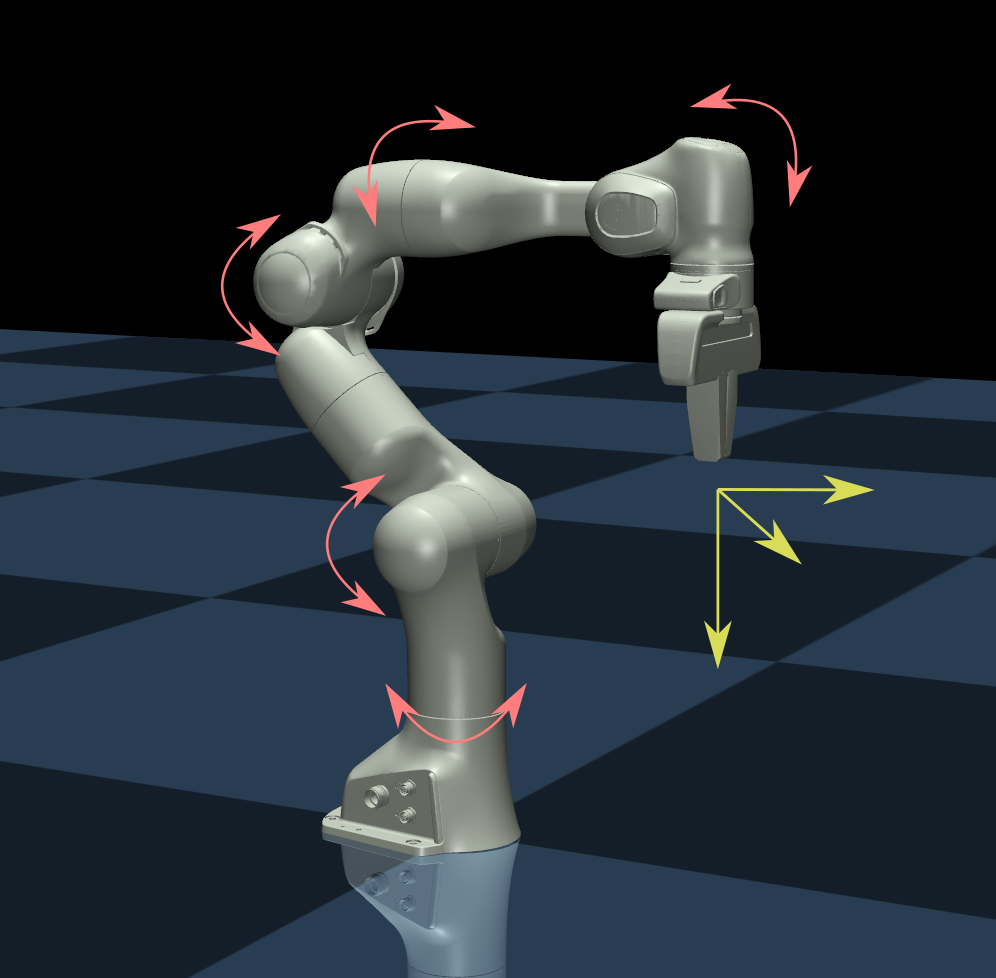}
    \captionof{figure}{Learning a policy for robot joint torques (red) to achieve an expert-demonstrated manipulation task is challenging due to the high dimensional configuration space. We propose to abstract the action space to the gripper pose movement (yellow) to allow imitation learning in a lower-dimensional action space.}
    \label{fig:toy_env}
\end{minipage}\hfill
\begin{minipage}{0.65\linewidth}
  \centering
  \includegraphics[width=\linewidth]{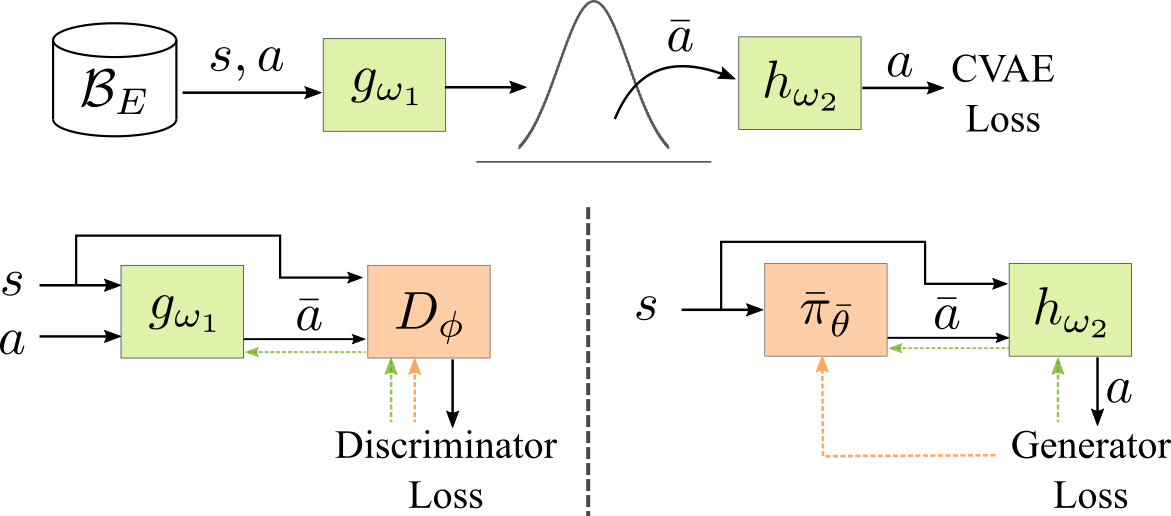}
  \captionof{figure}{\textbf{LAPAL overview}. We first train action encoder-decoder functions $g_{\omega_1}, h_{\omega_2}$ with a conditional variational autoencoder (CVAE) on expert demonstrations $\calB_E$ to extract latent action representation. In adversarial imitation learning, we iteratively train a discriminator $D_\phi$ that classifies state and latent action pairs $(\bfs, \bbfa)$ and train a generator $\bpi_\btheta$ that predicts latent actions from states. For task-agnostic LAPAL, we update the discriminator and generator parameters $\phi, \btheta$ by formulating a imitation learning objective in the latent space (orange dashed line). For task-aware LAPAL, the action encoder-decoder functions can be jointly optimized with the discriminator and policy (green dashed line).} 
  \label{fig:overview}
\end{minipage}
\end{table}

%% file: tex/RelatedWork.tex
\section{Related Work}
\label{sec:related_work}


IRL algorithms learn a policy by inferring a reward function from expert demonstration trajectories \citep{Neu2012Apprenticeship, Ng2000IRL, Abbeel2004IRL, Ratliff2006MMP, Ziebart2008MaxEnt}. Adversarial imitation learning (AIL) algorithms \citep{Ho2016GAIL, Finn2016GAN-GCL, Fu2017AIRL} establish a connection between IRL and GAN \citep{Goodfellow2014GAN}. AIL trains a discriminator that classifies between expert and agent transitions and a generator that based on a reward function defined through the discriminator. 

Existing methods on learning latent state representations are framed in a partially observable Markov decision process (POMDP) where latent belief states are learned to model the unobserved true states from visual observations. The observation log-likelihood is maximized with a variational lower bound. In RL, \citet{Hafner2019Planet} proposes to use a mixture of stochastic and deterministic sequential latent variable model and uses model predictive control to plan with learned dynamics and reward models. \citet{Lee2020SLAC} learns a purely stochastic sequential latent model from images and performs model-free reinforcement learning in the learned latent space. \citet{Gangwani2020BMIL, Rafailov2021VMAIL,Edwards2019ILPO} consider POMDPs for IRL and learn a latent belief state from visual observations for AIL algorithms. Our work focuses on a fully-observable Markov decision process (MDP) but it is challenging to directly apply AIL due to high dimensionality. We propose to learn a latent policy space to discover task structure and to enable efficient AIL training. In addition, \citet{Peng2018VDB} introduces a variational information bottleneck constraint when training the discriminator but they do not learn a transferrable latent policy. 

Learning latent action representations is studied in RL \citep{Allshire2021Laser} and offline RL \citep{Zhou2020Plas, Fujimoto2019BCQ}. \citet{Allshire2021Laser} proposed a conditional variational encoder-decoder (CVAE) model to learn latent action space where generated latent actions are decoded before applying on the original robot state. \citet{Van2020Plannable} learns latent spaces with equivariant maps to show that optimal policies are equivalent but the application is limited to small discrete systems where value iteration can be applied. \citet{Mishra2017Prediction} uses a CVAE to sample action sequences when optimizing policy with a learned latent dynamics model. \citet{Wang2017Robust} uses VAE to learn latent variables to encode full demonstration trajectories such that the discriminator is conditioned on the latent variable. Their contribution for learning latent trajecotry embedding is to generalize across diverse skills while we propose to discover latent task structure such that AIL can be trained efficiently.


%% file: tex/Preliminaries.tex
\section{Preliminaries: Generative Adversarial Imitation Learning}
\label{sec:preliminaries}

Consider a continuous-space control task formulated as a discrete-time Markov decision process (MDP) $\calM = \crl{\calS, \calA, T, r, \gamma}$, where $\bfs \in \calS \subseteq \bbR^n$ is the state, $\bfa \in \calA \subseteq \bbR^m$ is the action, $T: \calS \times \calA \times \calS \rightarrow [0, 1]$ is an unknown transition function from state $\bfs$ to state $\bfs'$ under action $\bfa$, $r: \calS \times \calA \rightarrow \bbR$ is a reward function and $\gamma \in (0, 1)$ is a discount factor. In IRL, the reward function $r$ is not known to the agent. Instead, it infers a reward function from a set of expert demonstrations $\crl{(\bfs_t, \bfa_t)}_t$ sampled from an expert policy $\pi_E: \calS \times \calA \rightarrow [0,1]$.

We denote the state-action marginal of the trajectory distribution induced by a policy $\pi$ in $\calM$ as $\rho_\pi(\bfs, \bfa)$. The imitation learning objective can be formulated as minimizing a general $f$-divergence between the state-action occupancy measures under a parameterized agent policy $\pi_\theta$, and the expert policy $\pi_E$,
\begin{equation}
\label{eq:f_div}
\min_\theta D_f \brl{\rho_{\pi_\theta}(\bfs, \bfa) \mid\mid \rho_{\pi_E}(\bfs, \bfa)},
\end{equation}
where $D_f[p||q] := \int f(\frac{p(\bfs, \bfa)}{q(\bfs, \bfa)}) q(\bfs, \bfa)d\bfs d\bfa)$ measures the difference between two probability distributions $p, q \in \calP(\calS\times\calA)$ in the space of density functions over $\calS\times\calA$, using a convex generator function $f : [0,\infty) \mapsto (-\infty,\infty]$. For example, AIRL \citep{Fu2017AIRL} optimizes the Kullback-Leibler divergence ($f(x) = x\ln x$), and GAIL \citep{Ho2016GAIL} optimizes the Jensen-Shannon divergence ($f(x) = \frac{1}{2}(x-1)\ln x$) \citep{Ho2016GAIL, Ghasemipour2020fMAX}. In this work, we consider the Jensen-Shannon divergence but the development can be generalized to other types of divergence functions as well. GAIL iteratively trains a discriminator $D_\phi$, and a generator policy $\pi_\theta$, with the following mini-max objective \citep{Ho2016GAIL, Ghasemipour2020fMAX}:
\begin{align}
\label{eq:minimax_objective}
\min_\theta\max_{\phi} J(\theta, \phi) &= \min_\theta D_{JS}\brl{\rho_{\pi_\theta}(\bfs, \bfa) \mid\mid \rho_{\pi_E}(\bfs, \bfa)} \notag \\
&= \min_{\theta}\max_{\phi} \bbE_{\prl{\bfs, \bfa}\sim\pi_E} \brl{\log D_\phi(\bfs, \bfa)} + \bbE_{\prl{\bfs, \bfa}\sim\pi_\theta} \brl{\log (1-D_\phi(\bfs, \bfa))}
\end{align}
where $D_\phi: \calS \times \calA \rightarrow [0, 1]$ is the discriminator function classifying the probability of a given state-action pair from the expert. In this paper, we will focus on solving \eqref{eq:minimax_objective} by introducing a latent policy defined on a learned latent action space. 

%% file: tex/LatentAction.tex
\section{Learning Latent Action Representation}
\label{sec:latent_action}

We consider lifting the original MDP to a new MDP, $\bar{\calM} = \crl{\calS, \bcalA, \bar{T}, \bar{r}, \gamma}$ where $\bcalA$ is a (lower-dimensional) latent action space, and $\bar{T}: \calS \times \bar{\calA} \times \calS \rightarrow [0, 1]$ and $\bar{r}: \calS \times \bar{\calA} \rightarrow \bbR$ are the transition probability and reward function defined in the latent space. We assume there exist optimal action encoder-decoder functions $g: \calA \rightarrow \bar{\calA}$ and $h: \bar{\calA} \rightarrow \calA$, such that $\bar{T}(\bfs, g(\bfa)) = T(\bfs, \bfa), \bar{r}(\bfs, g(\bfa)) = r(\bfs, \bfa), \forall \bfs \in \calS, \bfa \in \calA$. We further assume that the expert policy support lies within the image of the latent action decoder, i.e., $supp(\pi_E) \subset h(\bar{\calA}) \subset \calA$. The insight is that the expert policy only operates on a subset of the original configuration space. Therefore, learning a policy on such subspace is more efficient and it prevents explorations on the original action space where the optimal policy would never visit. In this section, we present a method to learn the latent action representations from expert demonstrations. In the following sections, we show that the learned action encoder-decoder functions can be fixed and \eqref{eq:minimax_objective} reduces to solving an imitation learning problem in latent MDP $\bcalM$, or they can be merged within the adversarial learning algorithm with their parameters optimized by gradients of \eqref{eq:minimax_objective}.


Following the motivation in Figure~\ref{fig:toy_env}, we propose to learn a latent action space such as the space of the robot's end-effector pose movements from it orginal action space of joint torques. To determine a latent action, the end-effector pose movement not only depends on the joint torques applied but also the current robot joint configuration state. Similarly, finding the joint torques such that the end-effector achieves the desired movement is also conditioned on the robot joint state. Inspired by these observations, we propose to use a conditional variational autoencoder (CVAE) \citep{Sohn2015CVAE} to model the action encoder-decoder functions and learn a latent represetation of the original action space. The action encoder function $g_{\omega_1}$ maps an action $\bfa$ to a latent distribution $E(\bbfa \mid \bfs)$, e.g. diagonal Gaussian, conditioned on the state $\bfs$, and samples a latent action $\bbfa$ from $E$. The action decoder function $h_{\omega_2}: \calS \times \bcalA \rightarrow \calA$ maps a latent action to the original action space, conditioned on the state $\bfs$. The loss function to train CVAE is 
\begin{equation}
\label{eq:cvae_loss}
\calL_{CVAE}(\bfs, \bfa) = ||\bfa - h_{\omega_2}(\bfs, g_{\omega_1}(\bfs, \bfa))||^2_2 + \beta D_{KL}[E(\bbfa \mid \bfs) \mid\mid p(\bbfa \mid \bfs)]
\end{equation}
where we assume a state-independent prior $p(\bbfa \mid \bfs) = p(\bbfa)$. The first reconstruction loss term ensures that we learn a latent action space that is consistent with the original action space and the second KL loss term regularizes the encoded latent posterior distribution. The coefficient $\beta$ is an adjustable hyperparameter that balances latent model capacity and reconstruction accurary \citep{Higgins2016betaVAE}. We train the CVAE model with expert demonstrations only so that the learned action space captures latent representations of the expert policy and not those of any random exploration policy for the reasons described in the previous paragraph.   

%% file: tex/TaskAgnostic.tex
\section{Task-agnostic Action Embedding for Adversarial Imitation Learning}
\label{sec:task_agnostic}

In this section, we assume the action encoder-decoder functions $g_{\omega_1}, h_{\omega_2}$ are trained as in Sec.~\ref{sec:latent_action} and the parameters $\omega_1, \omega_2$ are fixed. Given $g_{\omega_1}, h_{\omega_2}$, we can induce a latent MDP $\bar{\calM}$ as described in Sec.~\ref{sec:latent_action}. We consider an imitation learning problem in the latent MDP $\bar{\calM}$ and the mini-max objective anagolous to \eqref{eq:minimax_objective} is
\begin{align}
\label{eq:task_agnostic_minmax_objective}
\min_\btheta\max_{\phi} J(\btheta, \phi) &= \min_\btheta D_{JS}\brl{\rho_{\bpi_\btheta}(\bfs, \bbfa) \mid\mid \rho_{\bpi_E}(\bfs, \bbfa)} \notag \\
&= \min_\btheta\max_{\phi} \bbE_{\prl{\bfs, \bbfa}\sim\bpi_E} \brl{\log D_\phi(\bfs, \bbfa)} + \bbE_{\prl{\bfs, \bbfa}\sim\bpi_\btheta} \brl{\log (1-D_\phi(\bfs, \bbfa))}
\end{align}
Here, we abuse the notation $\prl{\bfs, \bbfa}\sim\bpi_E$ to denote that we have converted an expert transition $\prl{\bfs, \bfa}\sim\pi_E$ into the latent action space via $\bbfa = g_{\omega_1}(\bfs,\bfa)$. The latent agent policy $\bpi_\btheta: \calS \times \bcalA \rightarrow [0, 1]$ predicts a latent action $\bbfa$ at state $\bfs$, which is converted back to the original action space, $\bfa = h_{\omega_2}(\bfs,\bbfa)$, before applying it to the agent.

For discriminator optimization, we compute the gradient of the objective with respect to discriminator parameters $\phi$ as 
\begin{equation}
\label{eq:task_agnostic_discriminator_gradient}
\nabla_\phi J(\btheta, \phi) = \bbE_{\prl{\bfs, \bbfa}\sim\bpi_E} \brl{\nabla_\phi \log D_\phi(\bfs, \bar{\bfa})} + \bbE_{\prl{\bfs, \bbfa}\sim\bar{\pi}_\btheta} \brl{\nabla_\phi \log (1-D_\phi(\bfs, \bbfa))}.
\end{equation}
We can exchange the expectation with differentiation since the expectation terms do not depend on $\phi$. The expert latent actions are converted from the original state-action pair via the action encoder $\bbfa = g_{\omega_1}(\bfs, \bfa)$ and the agent latent action is sampled from the latent policy $\bpi$.

For generator optimization, we denote the discriminator $D_\phi$ evaluated at the optimal parameter $\phi^*$ as $D^* := D_{\phi^*}$. Since $\btheta$ appears in the probability distribution and not inside the expectation in the second term of \eqref{eq:task_agnostic_minmax_objective}, a gradient estimator can be obtained from the policy gradient theorem \citep{Sutton1999PG, Schulman2015Gradient},
\begin{align}
\label{eq:task_agnostic_generator_gradient}
\nabla_\btheta D_{JS}\brl{\rho_{\bpi_\btheta}(\bfs, \bbfa) \mid\mid \rho_{\bpi_E}(\bfs, \bbfa)} &= \nabla_\btheta \bbE_{\prl{\bfs, \bbfa}\sim\bar{\pi}_\btheta} \brl{\log (1-D^*(\bfs, \bbfa))} \notag \\
&= \bbE_{\prl{\bfs, \bbfa}\sim\bpi_\btheta} [\nabla_\btheta \log \bpi_\btheta (\bbfa\mid\bfs) Q^\bpi(\bfs, \bbfa)]
\end{align}
where $Q^\bpi(\bfs, \bbfa) = \bbE_{\prl{\bfs_t, \bbfa_t}\sim\bar{\pi}_\btheta}[\sum_{t=0}^\infty \gamma^t \log(1-D^*(\bfs_t, \bbfa_t)) \mid \bfs_0=\bfs, \bbfa_0=\bbfa]$ is the value function of $\bpi$ starting from $(\bfs, \bbfa)$. Effectively, the policy optimization step is to apply on-policy reinforcement learning with reward $\bar{r}(\bfs, \bbfa) = -\log(1 - D^*(\bfs, \bbfa))$ using on-policy samples from $\bpi_\btheta$. In practice, off-policy reinforcement learning algorithms, e.g., soft actor-critic (SAC) \citep{Haarnoja2018SAC}, can be applied to replace the expectation under the policy distribution with expectation under the agent replay buffer distribution \citep{Kostrikov2018DAC, Blonde2019SAM}. It is no longer guaranteed that the agent visitation distribution will match that of the expert but allows off-policy training for sample efficiency. In addition, since we optimize the discriminator and generator iteratively, we do not obtain the optimal discriminator $D^*$ but rather use the reward $\bar{r}(\bfs, \bbfa) = -\log(1 - D_\phi(\bfs, \bbfa))$ defined over the discriminator parameters in the current iteration. 

To sum up, in task-agnostic imitation learning, we assume that the action encoder-decoder functions $g_{\omega_1}, h_{\omega_2}$ are trained and their parameters are fixed. We apply GAIL in the latent MDP $\bar{\calM}$ where the discriminator $D_\phi$ classifies $(\bfs, \bbfa)$, and the policy $\bpi_\theta$ is trained with reward $\bar{r}(\bfs, \bbfa) = -\log(1 - D_\phi(\bfs, \bbfa))$ using SAC.

%% file: tex/TaskAware.tex
\section{Task-aware Action Embedding for Adversarial Imitation Learning}
\label{sec:task_aware}
In this section, we consider the case where the action encoder-decoder parameters $\omega_1, \omega_2$ are trainable, and solve the imitation learning optimization problem for the original MDP $\calM$. We rewrite the minimax objective in \eqref{eq:minimax_objective} as
\begin{align}
\label{eq:task_aware_minimax_objective}
\min_{\btheta, \omega_2}\max_{\phi, \omega_1} J(\btheta, \phi, \omega_1, \omega_2) &= \min_{\btheta, \omega_2}\max_{\phi, \omega_1} \bbE_{\prl{\bfs, \bfa}\sim\pi_E} \brl{\log D_\phi(\bfs, g_{\omega_1}(\bfs, \bfa))} \notag \\
& \;\;\;\;+ \bbE_{\prl{\bfs, \bfa}\sim\pi_{\btheta, \omega_2}} \brl{\log (1-D_\phi(\bfs, g_{\omega_1}(\bfs, \bfa)))}
\end{align}
where we have written the latent action as $\bbfa = g_{\omega_1}(\bfs, \bfa)$ and the latent policy samples $\prl{\bfs, \bbfa}\sim\pi_{\btheta}$ followed by action decoding $\bfa = h_{\omega_2}(\bfs,\bbfa)$ as $\prl{\bfs, \bfa}\sim\pi_{\btheta, \omega_2}$ to explicitly bring out the dependency on the parameters $\omega_1, \omega_2$.

For the discriminator update, we now compute the gradient with respect to both $\phi$ and $\omega_1$ since they affect the classification probability of a state-action pair $(\bfs, \bfa)$,
\begin{align}
\label{eq:task_aware_discriminator_gradient}
\nabla_{\phi, \omega_1} J(\btheta, \phi, \omega_1, \omega_2) &= \bbE_{\prl{\bfs, \bfa}\sim\pi_E} \brl{\nabla_{\phi, \omega_1} \log D_\phi(\bfs, g_{\omega_1}(\bfs, \bfa))} \notag \\
&\;\;\;\;+ \bbE_{\prl{\bfs, \bfa}\sim\pi_{\btheta, \omega_2}} \brl{\nabla_{\phi, \omega_1} \log (1-D_\phi(\bfs, g_{\omega_1}(\bfs, \bfa)))}.
\end{align}

For the generator update, we again notice that the policy parameters $\btheta, \omega_2$ only appear in the probability distribution and not inside the expectation. The policy gradient theorem can still be applied to obtain a gradient estimator,
\begin{align}
\label{eq:task_aware_generator_gradient}
\nabla_{\btheta, \omega_2} D_{JS}\brl{\rho_{\pi_{\btheta, \omega_2}}(\bfs, \bfa) \mid\mid \rho_{\pi_E}(\bfs, \bfa)} &= \nabla_{\btheta, \omega_2} \bbE_{\prl{\bfs, \bfa}\sim\pi_{\btheta, \omega_2}} \brl{\log (1-D^*(\bfs, g_{\omega_1}(\bfs, \bfa)))} \notag \\
&= \bbE_{\prl{\bfs, \bfa}\sim\pi_{\btheta, \omega_2}} [\nabla_{\btheta, \omega_2} \log \pi_{\btheta, \omega_2} (\bfa\mid\bfs) Q^\pi(\bfs, \bfa)],
\end{align}
where $Q^\pi_{\btheta, \omega_2}(\bfs, \bfa) = \bbE_{\prl{\bfs_t, \bfa_t}\sim\pi_{\btheta, \omega_2}}[\sum_{t=0}^\infty \gamma^t \log(1-D^*(\bfs_t, g_{\omega_1}(\bfs_t, \bfa_t))) \mid \bfs_0=\bfs, \bfa_0=\bfa]$ is the value function of $\pi_{\btheta, \omega_2} = h_{\omega_2} \circ \bpi_\btheta$ starting from $(\bfs, \bfa)$. In practice, we can optimize the agent policy $\pi_{\btheta, \omega_2}$ using any off-policy reinforcement learning algorithm, e.g. SAC, with reward function $r(\bfs, \bfa) = -\log(1 - D_\phi(\bfs, g_{\omega_1}(\bfs, \bfa))$ as described in the previous section. Our complete algorithm for task-agnostic and task-aware LAPAL is summarized in Algorithm~\ref{alg:LAPAL}.

\begin{algorithm}
\caption{Latent Policies for Adversarial Imitation Learning (LAPAL)}
\label{alg:LAPAL}
\begin{algorithmic}[1]
\State \textbf{Input}: Expert demonstration buffer $\calB_E$
\State Randomly initialize action encoder-decoder $\crl{g_{\omega_1}, h_{\omega_2}}$, discriminator $D_\phi$, latent policy $\bpi_\btheta$, and empty agent transition buffer $\calB_\pi$
\State Train action encoder-decoder $g_{\omega_1}, h_{\omega_2}$ with CVAE loss \eqref{eq:cvae_loss} using expert transitions from $\calB_E$
\For{number of training iterations}
  \For{number of experience collection steps}
    \State Sample latent action $\bbfa_t \sim \bpi_\btheta(\cdot \mid \bfs_t)$
    \State Decode action $\bfa_t = h_{\omega_2}(\bfs_t, \bbfa_t)$ and step environment for next state $\bfs_{t+1}$
    \State Add agent transitions $(\bfs_t, \bfa_t)$ to agent replay buffer $\calB_\pi$
  \EndFor
  \For{number of adversarial training steps}
    \State Sample a minibatch of transitions $\calB$ from joint buffer $\calB_E \cup \calB_\pi$
    \State Encode latent actions $\bbfa = g_{\omega_1}(\bfs, \bfa)$ for $(\bfs, \bfa) \in \calB$
    \If{using task-agnostic LAPAL} 
      \State Update discriminator parameters $\phi$ with gradient \eqref{eq:task_agnostic_discriminator_gradient}
      \State Update generator parameters $\btheta$ with reward $\bar{r}(\bfs, \bbfa) = -\log(1 - D_\phi(\bfs, \bbfa))$\\
      \hspace*{5em} using SAC~\citep{Haarnoja2018SAC}
    \ElsIf{using task-aware LAPAL} 
      \State Update discriminator parameters $\crl{\phi, \omega_1}$ with gradient \eqref{eq:task_aware_discriminator_gradient} 
      \State Update generator parameters $\crl{\btheta, \omega_2}$ with reward $r(\bfs, \bfa) = -\log(1 - D_\phi(\bfs, \bbfa))$\\ 
      \hspace*{5em} using SAC~\citep{Haarnoja2018SAC}
    \EndIf
  \EndFor
\EndFor
\end{algorithmic}
\end{algorithm}

%% file: tex/Limitations.tex
\section{Limitations}
\label{sec:limitations}

In this section we present a limitation on the theoretical aspect of our method where the objective function that we optimize is a lower unbound to the original imitation learning algorithm objective. Given a fixed action embedding function $f: \calA \rightarrow \bar{\calA}$, we induce a latent MDP $\bcalM$. The imitation learning objective function in $\calM$ is given as $D_f \brl{\rho_{\pi_\theta}(\bfs, \bfa) \mid\mid \rho_{\pi_E}(\bfs, \bfa)}$ while that in $\bcalM$ is $D_f\brl{\rho_{\bpi_\btheta}(\bfs, \bbfa) \mid\mid \rho_{\bpi_E}(\bfs, \bbfa)}$. The data processing theorem states that the latter is a lower bound for the former,
\begin{equation}
D_f\brl{\rho_\bpi(\bfs, \bbfa) \mid\mid \rho_{\bpi_E}(\bfs, \bbfa)} \leq D_f \brl{\rho_\pi(\bfs, \bfa) \mid\mid \rho_{\pi_E}(\bfs, \bfa)}.
\end{equation}
In other words, any processing of the ground truth state-action $(\bfs, \bfa)$ makes it more difficult to determine whether it came from the expert or the agent policy. As a result, the latent policy $\bpi$ that we solve in $\bcalM$ can be suboptimal in $\calM$ after action decoding, $\pi = h \circ \bpi$. Empirically we find that our task-agnostic LAPAL model can achieve a suboptimal performance in low dimensional systems where the action dimension is already small and embedding action space might lose important information for imitating expert policy. For large dimensional systems, the learned latent policy still matches the expert performance.

%% file: tex/Experiments.tex
\begin{figure}
  \centering
  \includegraphics[width=\linewidth]{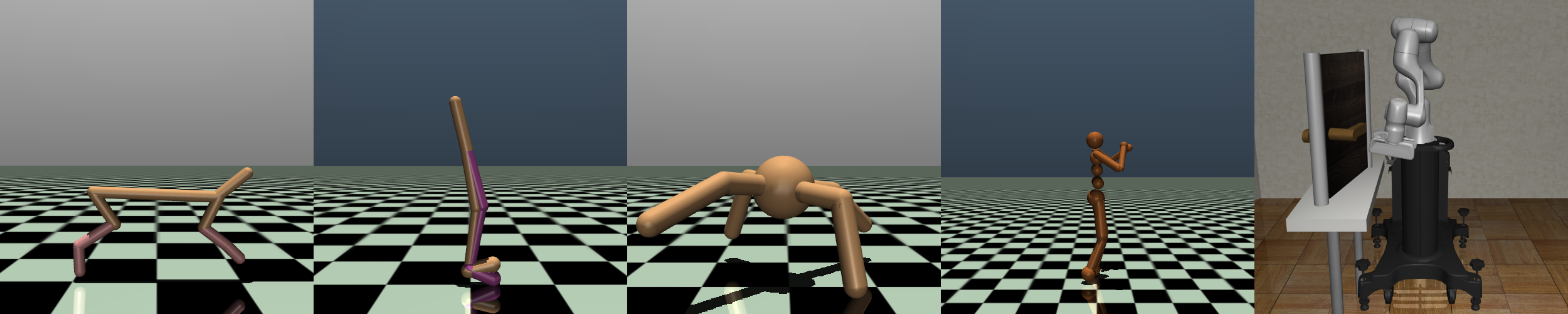}
  \caption{Benchmark environments from MuJoCo and robosuite: (left to right) HalfCheetah-v3, Walker2d-v3, Ant-v3, Humanoid-v3, Door.}
  \label{fig:benchmark_envs}
\end{figure}

\section{Experiments}
\label{sec:experiments}

\subsection{Benchmark tasks for imitation learning}

We use four continuous control locomotion environments from MuJoCo \citep{Todorov2012Mujoco, Brockman2016Gym} and one manipulation environment from robosuite \citep{robosuite2020}, presented in Fig.~\ref{fig:benchmark_envs}. For the locomotion environments, the task is to run at high speed without falling or exerting too much control effort. For the Door manipulation environment, the task is to open the door using a robot arm with a two-finger gripper. The original state and action dimensions are provided in Table~\ref{tb:env_specs}.

\begin{table}
\caption{Environment state and action space dimensions}
\label{tb:env_specs}
\centering
\begin{tabular}{ c  c  c } 
\Xhline{2\arrayrulewidth}
Environment & \makecell{State\\dimension} & \makecell{Action\\dimension} \\
\hline
HalfCheetah-v3 & 17 & 6 \\ 
Walker2d-v3 & 17 & 6 \\ 
Ant-v3 & 111 & 8 \\ 
Humanoid-v3 & 376 & 17 \\ 
Door (Panda/Sawyer) & 46 & 8 \\ 
\Xhline{2\arrayrulewidth}
\end{tabular}
\end{table}

We evaluate task-agnostic and task-aware LAPAL against GAIL \citep{Ho2016GAIL}. Each algorithm is provided with 64 expert demonstrations, collected from a policy trained with soft actor-critic (SAC) on the ground truth reward function. We use the same set of hyperparameters across GAIL and task-agnostic/aware LAPAL and the neural network architectures are shown in Table~\ref{tb:network_specs}. The generator policy SAC is adapted from \citet{Raffin2019SB3} and its critic and actor networks share weights for the first two layers. For LAPAL models, we add a Tanh activation layer to the action decoder output to match the action space bounds. The networks are trained in PyTorch \citep{Paszke2017PyTorch} with the Adam optimizer \citep{Kingma2014ADAM}. The latent action dimension is set to $4$ for task-agnostic/aware LAPAL to infer the latent task structure of these robotic locomotion and manipulation tasks.
\begin{table}
\caption{Neural network configurations}
\label{tb:network_specs}
\centering
\begin{tabular}{c c c c c} 
\Xhline{2\arrayrulewidth}
Module & Hidden layer size & Activation & Learning rate\\
\hline
Action encoder  & (256, 256) & Leaky ReLU & 3e-4 \\ 
Action decoder  & (256, 256) & Leaky ReLU & 3e-4 \\ 
Discriminator & (256, 256) & Tanh &  3e-5 \\ 
Generator actor & (256, 256, 256) & ReLU & 3e-4 \\ 
Generator critic & (256, 256) & ReLU & 3e-4 \\ 
\Xhline{2\arrayrulewidth}
\end{tabular}
\end{table}

\begin{figure}
  \centering
  \includegraphics[width=\linewidth]{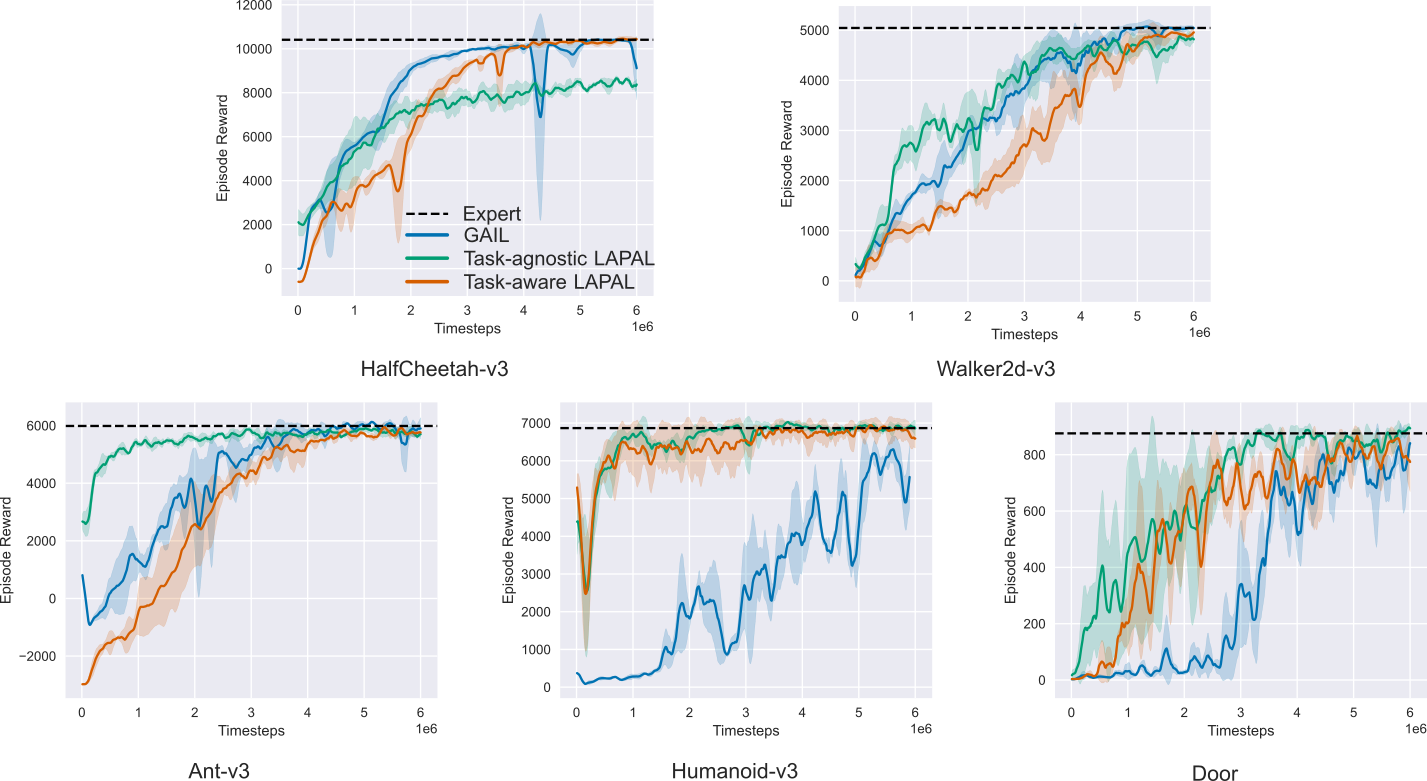}
  \caption{Benchmark results for MuJoCo and robosuite tasks. Each algorithm is averaged over 3 random seeds and the shaded area indicates standard deviation. LAPAL is on par with GAIL in low-dimensional problems such as Walker2d-v3 and HalfCheetah-v3 but converges faster for high-dimensional problems like Ant-v3, Humanoid and Door. In high-dimensional Humanoid-v3, GAIL fails to recover the optimal policy without addition regularization, e.g. gradient penalty and spectral normalization, while LAPAL converges quickly and asymptotically.}
  \label{fig:benchmark_results}
\end{figure}

Figure~\ref{fig:benchmark_results} shows the learning curves of each method. Task-agnostic LAPAL in HalfCheetah-v3 reaches a suboptimal performance and we conjecture that it is optimizing a lower bound of the imitation learning objective in the original space as discussed in Section~\ref{sec:limitations}. For high dimensional systems (Ant-v3, Humanoid, Door), both task-agnostic and task-aware LAPAL demonstrate faster convergence than GAIL. Specifically in the complex Humanoid-v3 environment, our models can achieve the expert baseline performance, while GAIL fails to recover an optimal policy in the original action space without explicit regularization techniques like gradient penalty \citep{Gulrajani2017WGAN, Orsini2021AIL} and spectral normalization \citep{Miyato2018Spectral}.

\subsection{Transfer learning in robosuite Door}
Our task-agnostic LAPAL model is suited for transfer learning problems in robotic tasks. We consider transferring skills acquired from one robot in a source environment $\calM_s$ to another robot in a task environment $\calM_t$. Specifically, we train task-agnostic LAPAL in the robosuite Door environment with a Panda robot arm. In the target Door environment, the robot is replaced with a Sawyer arm (see Figure~\ref{fig:transfer_envs}). The two robots have the same degrees of freedom but the joint configurations (geometry, friction, damping) and gripper models are different. Directly applying a policy trained from one robot to another for the same task does not work.

We first apply task-agnostic LAPAL in $\calM_s$ to acquire a latent policy $\bpi_s$ for the skill to open door. We then use $64$ expert demonstrations in $\calM_t$ to train a new action encoder-decoder $\crl{g_t, h_t}$ as described in Section~\ref{sec:latent_action} for the Sawyer arm. Combining the latent policy $\bpi_s$ in $\calM_s$ and the new action decoder $h_t$ we obtain the transferred policy for $\calM_t$. Table~\ref{tb:transfer_results} shows that without collecting additional samples from $\calM_t$, the transferred policy obtains an average return of $736$ while the expert policy for $\calM_t$ is $863$. This illustrates that LAPAL learns an informative latent policy for the manipulation task that can be generalized to different robot configurations. 

\begin{table}
\vspace*{-\baselineskip}
\begin{minipage}[t]{0.45\linewidth}
    \centering
    \strut\newline
    \includegraphics[width=\linewidth]{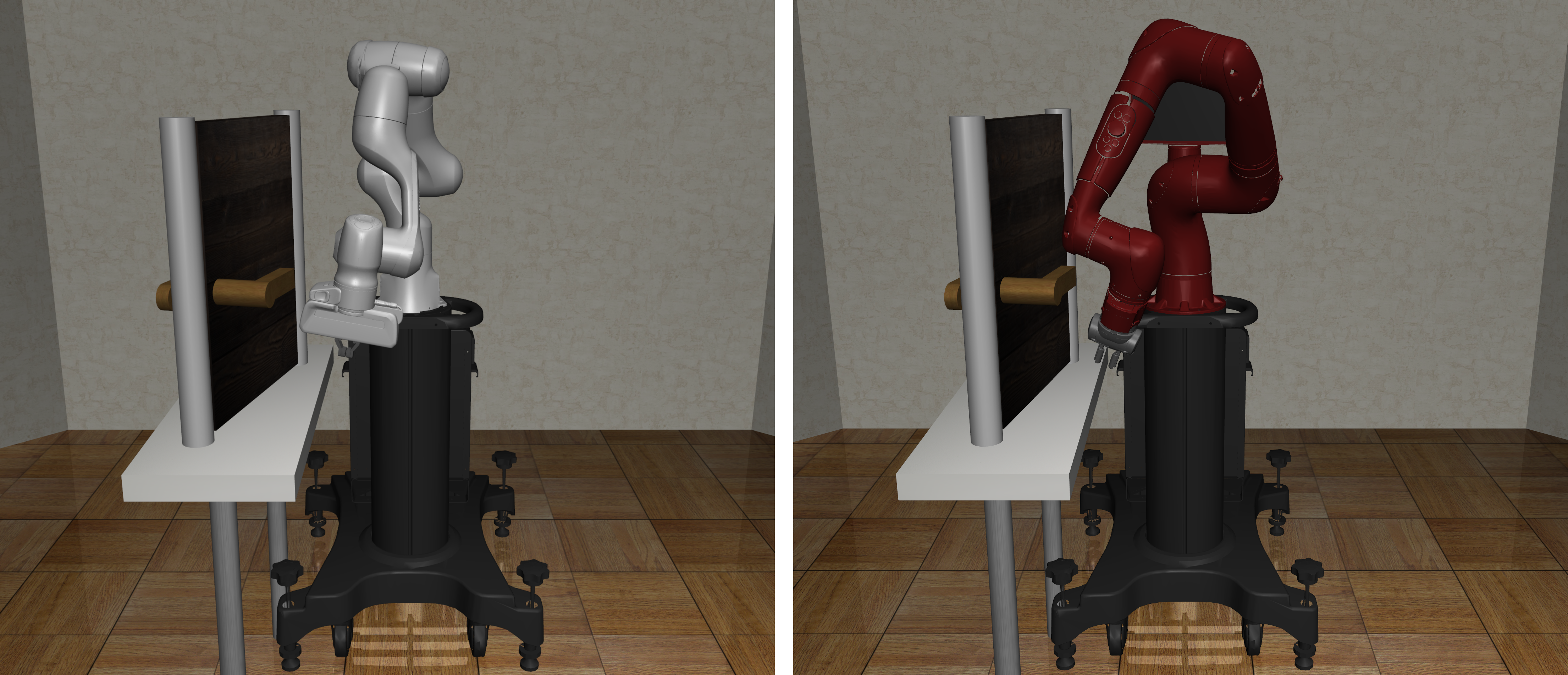}
    \captionof{figure}{Zero-shot transfer learning for task-agnostic LAPAL from Panda robot (left) to Sawyer robot (right) in Door environment.}
    \label{fig:transfer_envs}
\end{minipage}%
\hfill%
\begin{minipage}[t]{0.5\linewidth}
    \caption{Average return of each policy over 16 episodes in the Sawyer robot target environment $\calM_t$. The source policy is trained with a Panda robot in a source environment $\calM_s$ and directly applied to $\calM_t$. The transferred policy is a composition of the latent policy $\bpi_s$ trained in $\calM_s$ and the action decoder $h_t$ trained in $\calM_t$.} 
    \label{tb:transfer_results}
    \centering
    \begin{tabular}{ c c c c } 
     \Xhline{2\arrayrulewidth}
     & \makecell{Source\\policy} & \makecell{Transferred\\policy} & \makecell{Expert\\policy} \\
     \hline
     \makecell{Average\\return} & 33 & 736 & 863 \\ 
     \Xhline{2\arrayrulewidth}
    \end{tabular}
\end{minipage}
\end{table}

%% file: tex/Conclusion.tex
\section{Conclusion}
\label{sec:conclusion}

This paper introduced LAPAL, an approach that learns a latent action space to imitate expert behaviors efficiently in robotic locomotion and manipulation tasks. Our experiments show that LAPAL converges faster and yield significant improvements over a standard adversarial imitate learning baseline, especially in high-dimensional complex environments. Learning a latent state space in addition to an action space, such that the latent policy maps from latent states to latent actions, is an exciting future direction. State encoding may capture a common latent task structure, independent of the specific robot configuration, allowing a latent policy to generalize better for skill transfer across different robot types.